\pdfoutput=1
\documentclass[11pt]{article}
\usepackage{EMNLP2023}
\usepackage{times}
\usepackage{latexsym}

\usepackage[T1]{fontenc}

\usepackage[utf8]{inputenc}
\usepackage{microtype}

\usepackage{inconsolata}
\usepackage{microtype}
\usepackage{multirow}
\usepackage{graphicx}
\usepackage{amssymb}
\usepackage{wrapfig,booktabs}
\usepackage{eso-pic}

\newcommand{\emnlpfoottext}{%
  \parbox{\textwidth}{\centering\footnotesize
  \emph{Findings of the Association for Computational Linguistics: EMNLP 2024}, pages 173--180\\
  November 12--16, 2024 \copyright~2024 Association for Computational Linguistics
  }%
}

\AddToShipoutPictureFG{%
  \AtPageLowerLeft{%
    \ifnum\value{page}=1
      \raisebox{12mm}{
        \makebox[\paperwidth]{\hfill\emnlpfoottext\hfill}%
      }%
    \fi
  }%
}

\title{InsertGNN: A Hierarchical Graph Neural Network for the TOEFL Sentence Insertion Problem}
\author{Fang WU \\
  Stanford University, USA \\
  \texttt{fangwu97@stanford.edu} \\\And
  Stan Z. Li~\thanks{$\,\,\,$Corresponding author.} \\
  Westlake University, China \\
  \texttt{stan.zq.li@westlake.edu.cn} \\}

\begin{document}
\maketitle
\begin{abstract}
The integration of sentences poses an intriguing challenge within the realm of NLP, but it has not garnered the attention it deserves. Existing methods that focus on sentence arrangement, textual consistency, and question answering are inadequate in addressing this issue. To bridge this gap, we introduce InsertGNN, which conceptualizes the problem as a graph and employs a hierarchical Graph Neural Network (GNN) to comprehend the interplay between sentences. Our approach was rigorously evaluated on a TOEFL dataset, and its efficacy was further validated on the expansive arXiv dataset using cross-domain learning. Thorough experimentation unequivocally establishes InsertGNN's superiority over all comparative benchmarks, achieving an impressive 70\% accuracy—a performance on par with average human test scores.  
\end{abstract}

\section{Introduction and Related Work}
Sentence insertion (SI), initially introduced by~\citet{barzilay2008modeling}, stands as a crucial task for evaluating human linguistic prowess. However, with the advent of deep learning (DL), it has received scant attention over the past decade compared to other NLP domains like machine translation and text generation. As a result, the realm of DL lacks a standardized solution tailored to this particular challenge.
To address this void, we curate a dataset sourced from the TOEFL exam, an English proficiency test that archives test-takers' comprehensive accuracy scores. Through this dataset, we investigate whether contemporary NLP techniques can surpass the performance of nonnative English speakers~\footnote{Our compiled dataset is accessed at~\url{https://github.com/smiles724/TOEFL-Sentence-Insertion-Dataset}}

Sentence ordering (SO) and question answering (QA) constitute the two closely intertwined subdomains that share relevance with SI. On the one hand, SO initiates its approach by employing multi-layer perceptrons (MLPs) to facilitate pairwise order ranking~\citep{chen2016neural} but can inadvertently propagate errors to a significant degree. To mitigate this concern, the pointer network (PN)~\citep{gong2016end} integrates an attention mechanism to enhance model capacity, building on subsequent advances such as the incorporation of attention and the deepening of the architecture~\citep{logeswaran2016sentence,cui2018deep}. However, the direct applicability of the PN framework to SI is hindered by the differing nature of its input. As a remedy, recent studies solve SO by plugging a coherence verifier~\citep{jia2023sentence} or through a non-autoregressive manner~\citep{bin2023non}. Another line~\citep{putra2017evaluating} presents an unsupervised graph method that approaches the problem through the lens of sentence coherence and similarity within text. This method claims superiority when applied to the supervised entity grid and unsupervised Entity Graph scenarios. Yet, this approach, while effective, relies on a non-parametric structure and necessitates the construction of all potential graphs containing the extracted sentence. Meanwhile, QA provides a more universal pathway for our context. Notable works~\citep{joshi2020spanbert, abdel2023deep} have established QA as a potent framework. In our scenario, the removed sentence and the paragraph at large can be likened to the question and context, respectively. These components are seamlessly amalgamated and channeled through a network to produce the sought-after position. However, this linear fusion poses challenges for cutting-edge models such as Transformers in comprehending the inherent logic connecting the concatenated paragraph and its designated slots. 
\begin{figure*}[t]
\centering
\includegraphics[scale=0.55]{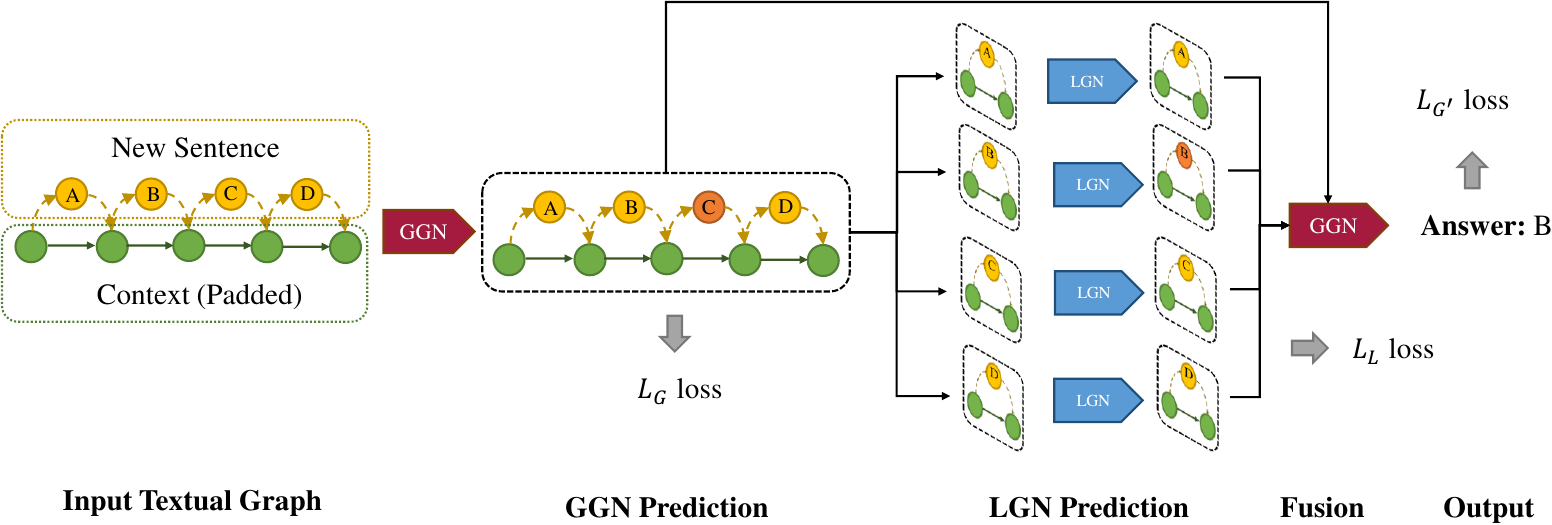}
\caption{The architecture of our proposed InsertGNN.}
\label{model}
\vspace{-1.5em}
\end{figure*}

In this paper, we represent SI as a directed graph, where each node represents a sentence, and the edge depicts their relative potential order. This pattern imitates the way people tackle SI, that is, putting the new sentence in each slot and checking the coherence. Then, we introduce a novel global-local fused GNN dubbed InsertGNN with delicately designed components to extract both global and local semantic information. Experiments on the TOEFL dataset with and without cross-domain learning convincingly show that InsertGNN surpasses all baselines and achieves competent performance compared to human examinees.  


\section{Methodology}
\subsection{Preliminary}
As the TOEFL SI question has four options ($A, B, C, D$), we divide the input paragraph into five parts as $\{c_i\}_{i=1}^5$ to accommodate this setting. Then we pad the paragraph to make the graph complete if there is no leading sentence before the slot $A$ or no ending sentence after the slot $D$. A graph $\mathcal{G}=(\mathcal{V},\mathcal{E})$ (see Fig.~\ref{model}) is built to describe all potential orders of the inserted sentence, where each node $v_i$ represents a sentence, and the directed edge $e_{ij}$ represents the relative order. If two sentences are connected, there is a directed edge between them. After that, we feed $\{c_i\}_{i=1}^5$ from the splitted context paragraph and the taken-out sentence $q$ into a sentence encoder, obtaining representation vectors $\{\mathbf{S}_{c_i}\}_{i=1}^5$ and $\mathbf{S}_q$. These vectors serve as node features, where $\mathbf{S}_q$ corresponds to features of node $\{A, B, C, D\}$.

\subsection{Global Graph Attention Networks}
Contextual semantics play a vital role in the determination of insertion positions. To begin with, we apply an $L$-layer Global Graph Attention Network (GGN)~\citep{velivckovic2017graph} to aggregate this global contextual information. The input feature is formulated as:
\begin{equation}
    \mathbf{H}^0=\{\underbrace{\mathbf{S}_{c1},...,\mathbf{S}_{c5}}_{\rm paragraph},\underbrace{\mathbf{S}_q,\mathbf{S}_q,\mathbf{S}_q,\mathbf{S}_q}_{\rm slots}\}. 
\end{equation}

Then for a center node $i$ and its neighbor $j$, the attention score at layer $l$ is computed as:
\begin{equation}
    e_{ij}=\rho(\alpha^T\cdot (\mathbf{W}\mathbf{H}_i^l \oplus \mathbf{W}\mathbf{H}_j^l)),
\end{equation}
where $\rho$ is the activation function and $\mathbf{W}$ is a trainable parameter. Attention weight is obtained by \emph{softmax} as $\alpha_{ij}=\frac{\exp{e_{ij}}}{\Sigma_{k\in N_i}{(\exp{e_{ik}}})}$. After that, $\alpha_{ij}$ are used to update node features as:
\begin{equation}
    \mathbf{H}^l_i=\sigma(\Sigma_{j\in N_i}\alpha_{ij}\mathbf{H}^{l-1}_j\mathbf{W}^l). 
\end{equation}

The representations for four slots in the last layer $\{\mathbf{H}_A^L,\mathbf{H}_B^L,\mathbf{H}_C^L,\mathbf{H}_D^L\}$ are fed into an MLP shared by the global-local fusion stage to obtain the prediction $\hat{y}$. Then, for a batch with $N$ samples, the binary cross-entropy (BCE) loss is calculated as:
\begin{equation}
    L_G=-\frac{1}{N}\sum_{i=0}^N[\hat{y}\log{y}+(1-\hat{y})\log{(1-y)}],
\end{equation}
where $y\in \{0, 1\}$ is the ground truth label.

\subsection{Local Graph Convolutional Networks}
\begin{figure}[t]
\centering
\includegraphics[scale=0.4]{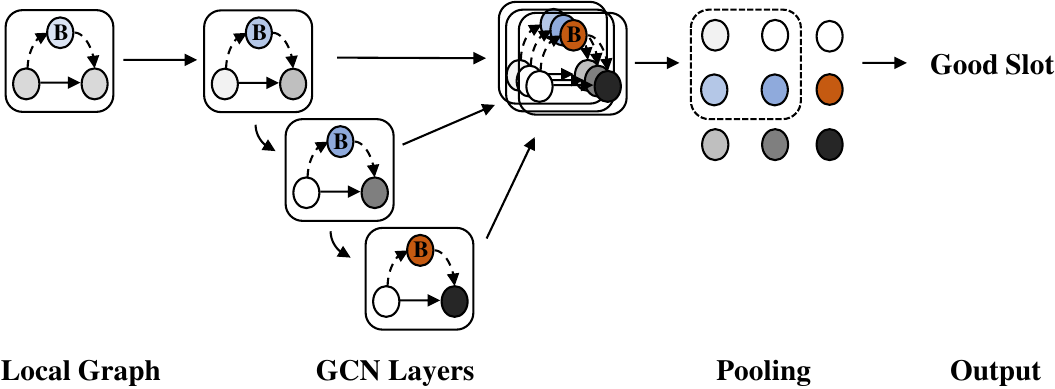}
\vspace{-0.5em}
\caption{The illustration of LGN layers..}
\label{lgn}
\vspace{-1.5em}
\end{figure}

However, GGN is insensitive to local details~\citep{zhang2020global}. Undeniably, the answer can sometimes be concluded by immediately reading the two sentences near the slot rather than the whole paragraph. Toward this goal, we utilize a Local Graph Network (LGN) to focus on local sentence interactions (see Fig.~\ref{lgn}). 

Concisely, we create four subgraphs with only the slot and its two surrounding sentences, whose features are the output of the previous GGN. Subgraphs are then fed into an $M$-layer parameter-shared GCN~\citep{kipf2016semi}, which is followed by a Weisfeiler-Lehman (WL) algorithm~\citep{weisfeiler1968reduction} to extract the multi-scale sub-tree features. The output of each layer $\mathbf{Z}^m$ is treated as WL's fingerprint. Similarly to DGCNN~\citep{phan2018dgcnn}, we horizontally concatenate these fingerprints $\{\mathbf{Z}^1,...,\mathbf{Z}^M\}$ rather than calculating the WL graph kernel. Then we apply a $4\times4$ pooling along with a fully-connected layer for subgraph classification. Finally, we compute the BCE loss $L_L$ for these four subgraphs. 

\subsection{Global-local Fusion}
At the fusion stage, $\mathbf{H}^L$ and $\mathbf{Z}^M$ are combined to integrate global and local information. We take a mean value of $\mathbf{Z}^M$ if node $i$ is contained in more than one subgraph. The fused features $\mathbf{E}=\mathbf{H}^L+ {\rm mean}(\mathbf{Z}^M)$
go through another GGN, and the output for four slots is fed into the shared MLP to attain the final prediction $\hat{y}'$, which is used during the inference period. After that, a BCE loss $L_{G'}$ is calculated. The total loss is the sum of three BCE losses as $L=\alpha L_G+\beta L_L+\gamma L_{G'}$, where $\alpha$, $\beta$, and $\gamma$ are loss weights of $L_G$, $L_L$, and $L_{G'}$.

\section{Experiment}
\subsection{Configurations and Datasets}
\paragraph{Setups.} We use the Sentence Transformer~\citep{reimers2019sentence} as the encoder to summarize the content of sentences, which makes sentences with similar meanings closer in vector space. It is first trained on Natural Language Inference (NLI) and then fine-tuned on the Semantic Textual Similarity benchmark (STSb) train set. Furthermore, we use BERT~\citep{devlin2018bert} and its two variants DistillBERT~\citep{sanh2019distilbert} and RoBERTa~\citep{liu2019roberta} for QA architecture. Notably, we neither fine-tune the baseline Transformers nor the Sentence Transformer, and only use them as an embedding layer. More details are in Appendix~\ref{exp_details}.

\paragraph{TOEFL exams.} TOEFL is one of the largest exams to test the English level of nonnative speakers, hosted by the Educational Testing Service (ETS) globally. We chose it for two main reasons. First, all TOEFL questions are extracted from academic articles, designed by language experts, and therefore are of high quality. Second, ETS annually offers score data reports of examinees. According to the latest summary, the average precision in the reading section is 70.67\%. 

Every year, ETS releases merely a few articles in TOEFL Practice Online, which prevents us from building a large-scale dataset. We collected all questions since 2011 and got 156 samples with a relatively equal label distribution of 32\%, 25\%, 22\%, and 21\%. 

\textbf{ArXiv abstracts.} We construct another dataset from arXiv to enrich the training samples and choose the abstract as the contextual paragraph, since it is independently readable and well-edited with strong logic clues. We abandon abstracts containing fewer than 5 sentences or 300 words to keep them more informative. Besides, we partially abandon categories that are not in the TOEFL scope. Moreover, several categories have tremendous mathematical formulations. These terms have no meaningful corresponding pretrained embeddings and should not be included in our supplementary dataset. 5965 abstracts remain after selection. 

NLTK~\citep{loper2002nltk} is used to break the paragraph into sentences and randomly choose one as the sentence to be inserted. Then, three other positions are selected to form a TOEFL-like problem. This operation can be repeated multiple times for each abstract since there are dozens of nonredundant combinations. The key statistics of these two datasets are listed in Appendix~\ref{detail_dataset}.

\subsection{Baselines}
\textbf{Unsupervised text coherence model.}~\citet{putra2017evaluating} propose three algorithms to build coherence graphs. The main difference is the determination of the edges (see Appendix~\ref{appendix_unsupervise}).

In the Preceding Adjacent Vertex (PAV), a weighted directed edge is established from each sentence to the preceding adjacent sentence. Single Similarity Vertex (SSV) discards the constraints of precedence and adjacency. Multiple Similar Vertex (MSV) even relaxes the singular condition and allows multiple outgoing edges for each sentence as long as their corresponding similarity score exceeds a threshold $\theta$. In the experiment, we use the same sentence encoder as our InsertGNN for the graph instead of GloVe~\citep{pennington2014glove}.

\textbf{Topological sort.} A research line~\citep{prabhumoye2020topological,sun2023japanese} regard SI as a constraint learning problem. Sentences between two slots are represented as nodes with a known constraint between them. An MLP is used to predict the remaining constraints of the relative order between the taken-out sentence and other sentences. {~\citet{lai2021improving} extend Sentence-Entity Graph Recurrent Network (SE-GRN)~\citep{yin2019graph} and utilize two graph-based classifiers to iteratively make pairwise predictions for pairwise sentences.}

\textbf{QA methods.} Here, we formulate two types of QA architecture. The \emph{P}-type (Plain) linearly combines the dependent sentence and the paragraph as the input and extracts the output vector of the [CLS] character as the final representation. The \emph{A}-type (Altered) puts the new sentence in all four possible slots and classifies those differently filled paragraphs. The final prediction will be the one with the highest probability.

\textbf{Large language models (LLMs).} LLMs~\citep{zhao2023survey} are posing a significant impact on the AI community. Here, we evaluate ChatGPT-3.5 with different prompts and report the best one. More details are listed in Appendix~\ref{app_chatgpt}.

\subsection{Results}
\paragraph{TOEFL.} We test the unsupervised text coherence model first. PAV attains the highest accuracy of 41.66$\%$ on our TOEFL dataset (see Table~\ref {unsupervised performance}), in accord with~\citet{putra2017evaluating}’s evaluation. It indicates that local cohesion is more important than long-distance cohesion, in line with our motivation for LGNs. 
\begin{table}[t]
\centering
\resizebox{0.55 \columnwidth}{!}{%
\begin{tabular}{cc}\toprule
    \textbf{Method} & \textbf{Acc\_TOEFL (\%)} \\ \midrule
    PAV & \textbf{41.66} \\
    SSV & 34.62 \\
    MSV & \underline{37.18} \\ \bottomrule
\end{tabular}}
\vspace{-0.5em}
\caption{Unsupervised learning accuracy. The best and second best are in bold and underlined, respectively. }
\label{unsupervised performance}
\vspace{-1.5em}
\end{table}

Next, we test the performance of supervised approaches with different proportions of validation.  Experiments are repeated three times with different seeds, and we report the mean value. The result shows our InsertGNN observably improves upon other baselines, including ChatGPT-3.5 in all cases. To be specific, InsertGNN reaches the highest accuracy of 71.5$\%$ with abundant training samples (see Table~\ref {toefl performance}). It is comparable to the average human accuracy of 70.67$\%$, suggesting that our model can do at least as well as non-native English speakers. We also conducted an ablation study to examine the contribution of each loss, and it can be discovered that all components are non-redundant (see Appendix~\ref{appendix_ablation}).  

\begin{table}[t]
\centering
\resizebox{1.0 \columnwidth}{!}{%
\begin{tabular}{cccc} \toprule
    \multicolumn{2}{c}{\textbf{Method}} & \textbf{\begin{tabular}[c]{@{}c@{}}Acc\_TOEFL (\%) \\ (dev=0.05)\end{tabular}} & \textbf{\begin{tabular}[c]{@{}c@{}}Acc\_TOEFL (\%)  \\ (dev=0.5)\end{tabular}} \\ \midrule
    \multirow{2}{*}{BERT} & \emph{P} & 42.86 & 38.46 \\ 
     & \emph{A} & 42.86 & 32.05 \\ \midrule
    \multirow{2}{*}{DistillBERT} & \emph{P} & 57.14 & 35.89 \\ 
     & \emph{A} & 57.14 & 29.49 \\ \midrule
    \multirow{2}{*}{RoBERTa} & \emph{P} & 42.86 & 35.89 \\ 
     & \emph{A} & 42.86 & 28.61 \\ \midrule
    \multicolumn{2}{c}{Topological Sort} & 57.14 & 34.62 \\ 
    \multicolumn{2}{c}{SE-GRN} & {57.14} & {46.15} \\ 
    \multicolumn{2}{c}{ChatGPT-3.5} & \underline{61.52} & \underline{53.84}  \\ \midrule
    \multicolumn{2}{c}{InsertGNN} & \textbf{71.43} & \textbf{55.12} \\  \bottomrule
\end{tabular}}
\caption{Supervised learning accuracy under different validation split ratios of 0.05 and 0.5. \emph{P} and \emph{A} refers to the \emph{P}-type and the \emph{A}-type QA structure.}
\label{toefl performance}
\end{table}

\paragraph{From ArXiv to TOEFL.} We further evaluate InsertGNN using cross-domain learning. Specifically, models are first trained on the arXiv dataset (source domain) with a validation splitting ratio of 0.05 and then directly tested on the TOEFL dataset (target domain). InsertGNN still stands out with an accuracy of 39.1$\%$ (see Table~\ref {arxiv}), outperforming all baselines except ChatGPT-3.5. 
\begin{table}[t]
\centering
\resizebox{1.0 \columnwidth}{!}{%
\begin{tabular}{cccc}\toprule
    \multicolumn{2}{c}{\textbf{Method}} & \textbf{Acc\_arXiv (\%)} & \textbf{Acc\_TOEFL (\%)} \\ \midrule
    \multirow{2}{*}{BERT} & \emph{P} & 34.74 & 33.97 \\ 
     & \emph{A} & 32.63 & 30.13 \\ \midrule
    \multirow{2}{*}{DistillBERT} & \emph{P} & 35.22 & 33.97 \\ 
     & \emph{A} & 31.33 & 30.76 \\ \midrule
    \multirow{2}{*}{RoBERTa} & \emph{P} & 33.56 & 32.69 \\ \cline{2-4} 
     & \emph{A} & 3255 & 3141 \\ \midrule
    \multicolumn{2}{c}{Topological Sort} & 43.62 & 28.85 \\ 
    \multicolumn{2}{c}{SE-GRN} & \underline{44.96} & {36.53} \\
    \multicolumn{2}{c}{ChatGPT-3.5} & -- & \textbf{60.76} \\ \midrule
    \multicolumn{2}{c}{InsertGNN} & \textbf{46.31} & \underline{39.10} \\ \bottomrule
\end{tabular}}
\caption{Cross-domain learning accuracy from the arXiv dataset to the TOEFL dataset. The left and right columns correspond to the accuracy in the arXiv test set and the whole TOEFL dataset, respectively.}
\label{arxiv}
\vspace{-1em}
\end{table}

Moreover, the accuracy in TOEFL is generally lower than in arXiv because the contents of the two datasets are slightly different. ArXiv abstracts are a brief summarization and, therefore, very condensed. In contrast, the TOEFL paragraphs are an expanded narrative of a sub-point or a detailed explanation, which is more elaborate with a stronger inner logic. This manner of writing causes a decrease in accuracy when models are cross-domain. 
\section{Discussion}
It is worth mentioning that the SI problem shares similarities with existing pretraining objectives like BERT's Next Sentence Prediction (NSP) and ALBERT's Sentence-Order Prediction (SOP). Specifically, NSP concatenates two masked sentences as inputs during pretraining. Sometimes they correspond to sentences that were next to each other in the original text, sometimes not. The model then has to predict if the two sentences were following each other or not. SOP primarily focuses on inter-sentence coherence and is designed to address the ineffectiveness of the next sentence prediction (NSP) loss proposed in the original BERT.

\section{Conclusion}   
In the paper, we propose a new sentence insertion framework called InsertGNN and build a TOEFL benchmark data set. Strong empirical evidence demonstrates the effectiveness of InsertGNN over existing language models like ChatGPT in this interesting NLP task.  

\section{Limitations and Future Work}
Despite the superior performance of our model, there are still some limitations left for future work. Most importantly, the TOEFL sentence insertion is small in size. Though we offer a compensatory dataset curated from the arXiv abstracts, more effort is needed to collect a larger and higher-quality SI dataset. In addition, more powerful GPT models such as ChatGPT-4.0 are emerging, and it is worth evaluating these more advanced tools in our TOEFL sentence insertion task. Also, we believe more effective prompts can be mined to solve this specific task, and we leave this for future study. 
\bibliography{cite}
\bibliographystyle{acl_natbib}

\appendix
\newpage
\onecolumn
\section{Details of Datasets}
\label{detail_dataset}
There we give an example of the SI problem in the TOEFL efficiency test (see Table~\ref{toefl}). 
\begin{table*}[h]
\centering
\resizebox{0.9 \columnwidth}{!}{%
\begin{tabular}{c|c} \toprule 
\textbf{Context Paragraph} & \begin{tabular}[c]{@{}l@{}} \textbf{{[}A{]}} The age of rock art in   Australia has been revised several times,\\  with earlier dates suggested recently after discoveries.\\  \textbf{{[}B{]} }Accurate scientific proof has dated the first appearance of\\  surface rock in Australia to approximately 30,000 to 50,000 years ago. \\ \textbf{{[}C{]}} This lengthy and astounding history of rock art in Australia\\  makes it the oldest art tradition known today in the world.   \textbf{{[}D{]}}\end{tabular} \\ \midrule
\textbf{New Sentence} & \begin{tabular}[c]{@{}l@{}}Thanks to radiocarbon dating and technological development in studying \\  evidence, researchers can now give a more precise age on this type of art.\end{tabular} \\ \midrule
\textbf{Answer} & \begin{tabular}[c]{@{}l@{}}\textbf{{[}B{]}} is the correct answer. Here, we need to match back “this type of art” \\to what it is referencing in order to correctly place the prompt sentence. \\It’s referencing rock art, named in the first sentence, making the answer \textbf{{[}B{]}}.\end{tabular} \\ \bottomrule
\end{tabular}}
\caption{An example of a TOEFL sentence insertion problem. It is extracted from questions of the third reading passage named "Rock Art of the Australian Aborigines" from TOEFL Practice Online 23.}
\label{toefl}
\end{table*}

In the experiments, we use two sorts of datasets: one is the collected TOEFL dataset, and the other is the arXiv abstract dataset. We summarize their statistics in Table~\ref{dataset}. 
\begin{table*}[h]
\centering
\resizebox{0.95 \columnwidth}{!}{%
\begin{tabular}{c|cccc} \toprule
    Dataset & {Size} & {Sentences} & {Words} & {Topics} \\ \midrule
    \textbf{TOEFL} & 156 & 7.31 & 133.94 & \begin{tabular}[c]{@{}c@{}}Anthropology, Architecture, Astronomy, Economics, Biology, \\ Chemistry, Communication, North America, Physics, \\ Political Science, Psychology Sociology, World History\end{tabular} \\ \midrule
    \textbf{arXiv} & 5965 & 7.24 & 121.36 & \begin{tabular}[c]{@{}c@{}}Astrophysics, Computer Vision and Pattern Recognition, \\ Cryptography, Economics, General Relativity\\ High Energy Physics-Theory, Information Theory,\\ Networking and Internet Architecture, Quantum Physics\end{tabular} \\ \bottomrule
\end{tabular}}
\caption{Dataset statistics, including the sample size, average number of sentences inside each paragraph, the average number of words of each paragraph, and their main topics (categories).}
\label{dataset}
\end{table*}

\section{Some Baseline Models}
\label{appendix_unsupervise}
\begin{figure}[ht]
\centering
\includegraphics[scale=0.35]{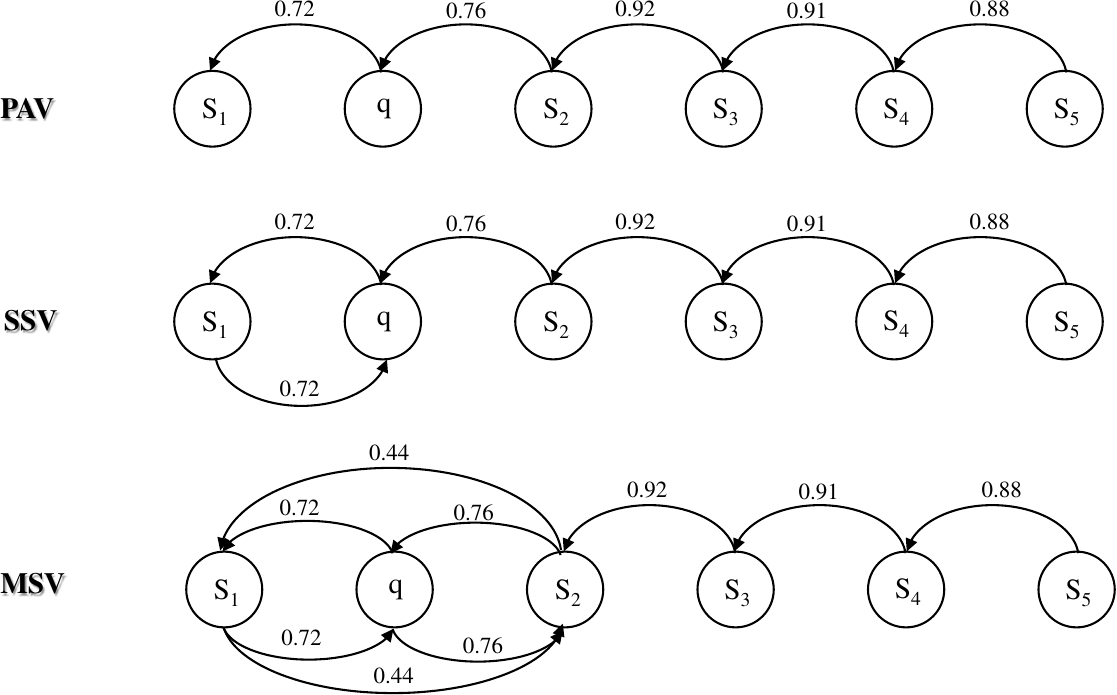}
\caption{Three graph construction algorithms. PAV only allows edges between a sentence and its preceding sentence. SSV allows for edges between a sentence and any other sentence. MSV allows multiple edges.}
\label{unsupervise}
\end{figure}

We provide a visualization of how the unsupervised text coherence method~\citep{putra2017evaluating} processes the SI problem in Fig.~\ref{unsupervise}. 

\section{Experimental Details}
\label{exp_details}
There, we explicitly describe the experimental configurations. All learnable models are trained on a single A100 GPU. Regarding MSV, we choose a threshold $\theta$ of 0.3. For both GGNs and LGNs, we utilize leaky Relu and Relu as the activation function, respectively, and include the dropout mechanism between layers with a dropout rate of 0.5. They both have one hidden layer and four hidden units. For GGN, we choose 16 hidden attention heads and four output attention heads with an attention dropout of 0.6 and a 0.2 negative slope of leaky Relu. They all have a residual connection. For all MLPs, we utilize Tanh as the activation function with no dropout. We adopt an Adam optimizer with a weight decay rate of 0.0005 and a random seed of 1234. After a grid search algorithm, we set $\alpha=0.2$, $\beta=0.2$, and $\gamma=1.0$, where $L_{G'}$ is given more weights so that the model can focus more on the global-local fusion information. We train 100 epochs for InsertGNN with a learning rate of 0.0001 and 200 epochs for BERT-based models with a learning rate of 0.01.

\subsection{Implementation of ChatGPT-3.5}
\label{app_chatgpt}
For the powerful toolkit ChatGPT~\footnote{\url{https://openai.com/blog/chatgpt/}}, prompt engineering is a key factor. We tried three different prompts to obtain the answer, which are listed below. We also provide the response template provided by ChatGPT-3.5. The empirical result shows that performance varies significantly according to different prompt inputs. The precision of prompts 1, 2, and 3 are 53.82\%, 61.52\%, and 30.76\%, respectively, and we can find that prompt 2 is optimal. 

\paragraph{Prompt 1.} Please do the following sentence insertion problem, which has four choices denoted as A/B/C/D. The context paragraph is \emph{xxx} and the sentence to be inserted is \emph{xxx}. What is the answer? 

\paragraph{Prompt 2.} Given a sentence \emph{xxx}, which place should it be inserted in the following paragraph \emph{xxx}?

\paragraph{Prompt 3.} Which paragraph is the most fluent? \emph{xxx}, \emph{xxx}, \emph{xxx}, or \emph{xxx}.

\paragraph{ChatGPT-3.5's answer for prompt 1.} The most appropriate insertion point for the given sentence is before sentence [A/B/C/D]. Here is the revised paragraph: \emph{xxx}.

\paragraph{ChatGPT-3.5's answer for prompt 2.} The sentence \emph{xxx} fits best after the sentence [A / B / C / D] in the paragraph. Here is the revised paragraph: \emph{xxx}.

\paragraph{ChatGPT-3.5's answer for prompt 3.} The most fluent paragraph is the first/second/third/fourth one: \emph{xxx}. 

\section{Additional Results}
\label{appendix_ablation}
We implement ablation studies in the TOEFL data set with a 0.5 validation split ratio to verify the contributions of each constituent of our InsertGNN in Table~\ref{ablation}. It can be observed that the LGN significantly promotes the model performance.  
\begin{table}[h]
\centering
\resizebox{0.3 \columnwidth}{!}{%
\begin{tabular}{c | ccc | c} \toprule
     & $L_G$ & $L_L$ & $L_{G'}$ &  Accuracy  \\ \midrule
    1 & \checkmark & -  & - & 0.4487 \\
    2 & \checkmark & \checkmark  & - & 0.5256 \\
    3 & \checkmark & \checkmark  & \checkmark & \textbf{0.5512} \\\bottomrule
\end{tabular}}
\caption{Ablation studies.}
\label{ablation}
\end{table}

\end{document}